\documentclass[a4paper]{article}
\usepackage[latin9]{inputenc}
\usepackage{array}
\usepackage{float}
\usepackage{multirow}
\usepackage{amsmath}
\usepackage{amssymb}
\usepackage{graphicx}

\makeatletter

\pdfpageheight\paperheight
\pdfpagewidth\paperwidth

\providecommand{\tabularnewline}{\\}

\usepackage{spconf}

\usepackage{bm}

\ninept

\title{
Multilingual sequence-to-sequence speech recognition:\\
Architecture, transfer learning, and language modeling
}
\name{\parbox{1.0\linewidth}{\center
Jaejin Cho$^{1,\ddagger}$ \thanks{${\ddagger}$  All three authors share equal contribution}, Murali Karthick Baskar$^{2,\ddagger}$, Ruizhi Li$^{1,\ddagger}$, Matthew Wiesner$^1$,
 \\ Sri Harish Mallidi$^3$, Nelson Yalta$^4$, Martin Karafi\'{a}t$^2$, Shinji Watanabe$^1$, Takaaki Hori$^5$}}

\address{
$^1$Johns Hopkins University, $^2$Brno University of Technology, $^3$Amazon, $^4$Waseda University, \\$^5$Mitsubishi Electric Research Laboratories (MERL)\\
{\small \tt \{ruizhili,jcho52,shinjiw\}@jhu.edu,\{baskar,karafiat\}@fit.vutbr.cz,thori@merl.com}
}

\@ifundefined{showcaptionsetup}{}{%
 \PassOptionsToPackage{caption=false}{subfig}}
\usepackage{subfig}
\makeatother

\begin{document}
\ninept\maketitle 
\begin{abstract}
\end{abstract}
Sequence-to-sequence (seq2seq) approach for low-resource ASR is a
relatively new direction in speech research. The approach benefits
by performing model training without using lexicon and alignments.
However, this poses a new problem of requiring more data compared to conventional
DNN-HMM systems. In this work, we attempt to use data from 10 BABEL
languages to build a multilingual seq2seq model as a prior model, and then port them towards 4 other BABEL languages using transfer learning approach. We also explore different architectures for improving the prior multilingual seq2seq model.
The paper also discusses the effect of integrating a recurrent neural network language model (RNNLM) with a seq2seq model during decoding.
Experimental results show
that the transfer learning approach from the multilingual model shows substantial gains over monolingual models across all 4 BABEL languages. 
Incorporating an RNNLM also brings significant improvements in terms of \%WER, and achieves recognition performance comparable to the models trained with twice more training data.


\section{Introduction}

The sequence-to-sequence (seq2seq) model proposed in \cite{sutskever2014sequence,bahdanau2014neural,cho2014learning}
is a neural architecture for performing sequence classification and later adopted to perform speech recognition in ~\cite{chorowski2015attention,graves2014towards,graves2012supervised}.
The model allows to integrate the main blocks of ASR such as acoustic
model, alignment model and language model into a single framework.
The recent ASR advancements in connectionist temporal classification
(CTC) \cite{graves2012supervised,graves2014towards} and attention
\cite{chorowski2015attention,chan2016listen} based approaches has created larger
interest in speech community to use seq2seq models. To leverage performance
gains from this model as similar or better to conventional hybrid
RNN/DNN-HMM models requires a huge amount of data \cite{rosenberg2017end}.
Intuitively, this is due to the wide-range role of the model in performing alignment
and language modeling along with acoustic to character label mapping
at each iteration.

In this paper, we explore the multilingual training approaches \cite{besacier2014automatic,tuske2014multilingual,karafiat2016multilingual} used in hybrid DNN/RNN-HMMs to incorporate them into the seq2seq models.
In a context of applications of multilingual approaches
towards seq2seq model, CTC is mainly used instead of the attention models. 
A multilingual CTC is proposed in \cite{tong2017multilingual}, which
uses a universal phoneset, FST decoder and language model. The authors
also use linear hidden unit contribution (LHUC) \cite{swietojanski2014learning}
technique to rescale the hidden unit outputs for each language as
a way to adapt to a particular language. 
Another work \cite{muller2017language} on multilingual CTC shows the importance of language adaptive vectors as auxiliary input to the encoder in multilingual CTC model. The decoder
used here is a simple $argmax$ decoder. An extensive analysis on
multilingual CTC mainly focusing on improving under limited data condition is performed in \cite{dalmia2018sequence}. 
Here, the authors use a word level FST decoder integrated with CTC during decoding.

On a similar front, attention models are explored within a multilingual
setup in \cite{watanabe2017language,toshniwal2017multilingual} based on attention-based seq2seq to build a
model from multiple languages. 
The data is just combined together assuming the target languages are seen during the training. And, hence
no special transfer learning techniques were used here to address
the unseen languages during training.
The main motivation and contribution behind this work is as follows:
\begin{itemize}
\item To incorporate the existing multilingual approaches in a joint CTC-attention~\cite{watanabe2017hybrid} (seq2seq) framework, which uses a simple beam-search decoder as described in sections \ref{sec:seq2seq} and \ref{sec:Multilingual-experiments}
\item Investigate the effectiveness of transferring a 
multilingual model to a target language under various data sizes. This is explained in section \ref{subsec:finetune}.
\item Tackle the low-resource data condition with both transfer learning and including a character-based RNNLM trained with multiple languages. Section \ref{subsec:Multilingual-RNNLM} explains this in detail.

\end {itemize}

\section{Sequence-to-Sequence Model}\label{sec:seq2seq}

\begin{figure}[tb]
\includegraphics[width=8.3cm]{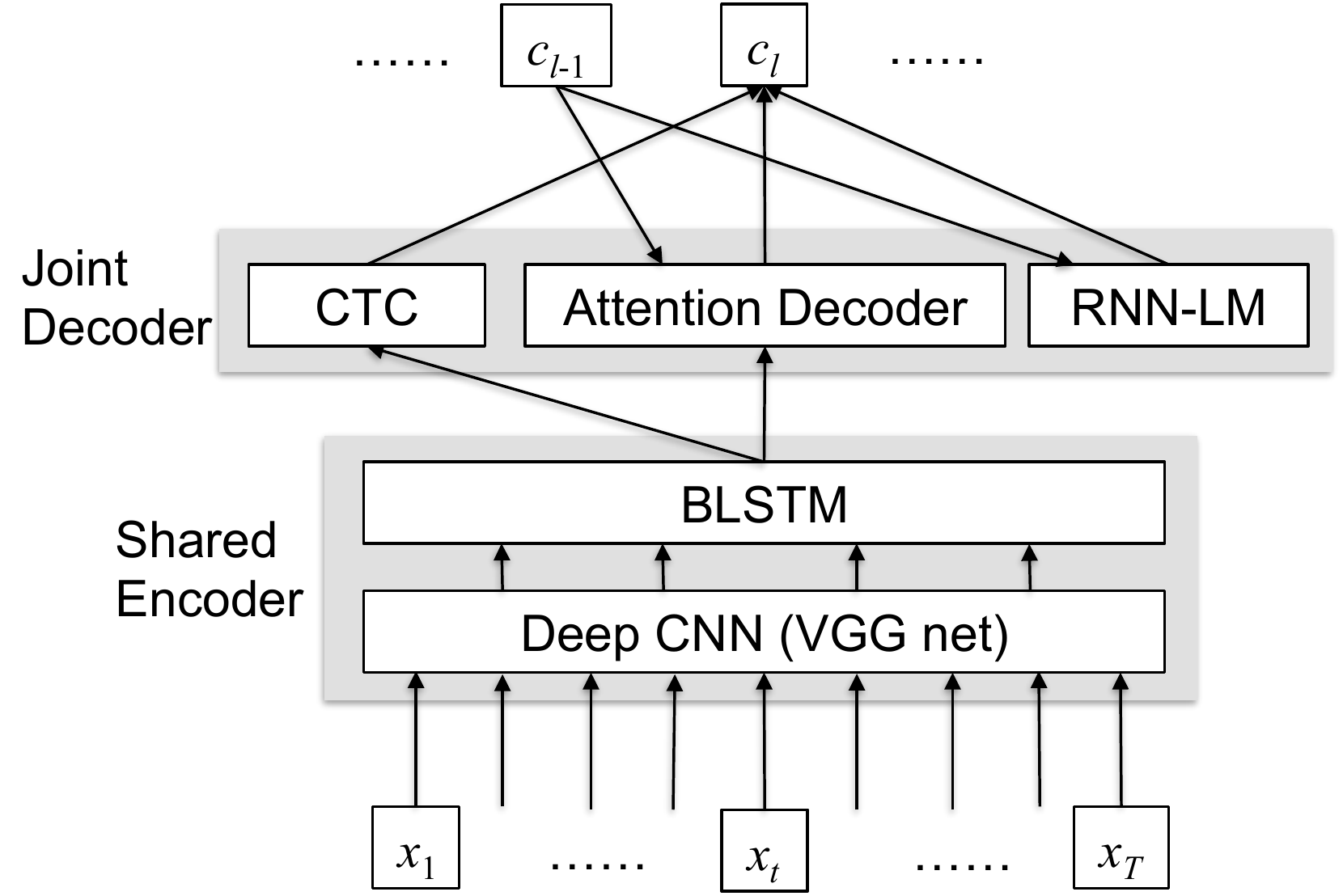}
\caption{Hybrid attention/CTC network with LM extension: the shared encoder is trained by both CTC and attention model objectives simultaneously. The joint decoder predicts an output label sequence by the CTC, attention decoder and RNN-LM.}
\label{fig:ctc-attention}
\end{figure}

In this work, we use the attention based approach \cite{bahdanau2014neural}
as it provides an effective methodology to perform sequence-to-sequence
(seq2seq) training. Considering the limitations of attention in performing
monotonic alignment \cite{sperber18interspeech,DBLP:journals/corr/abs-1712-05382},
we choose to use CTC loss function to aid the attention mechanism in both training and decoding. The basic network architecture is shown in Fig. \ref{fig:ctc-attention}.

Let $X = (\mathbf{x}_t | t=1, \dots, T)$ be a $T$-length speech feature sequence and $C= (c_l | l=1, \dots, L)$ be a $L$-length grapheme sequence.
A multi-objective learning framework $\mathcal{L}_{\text{mol}}$ proposed in \cite{watanabe2017hybrid} is used in this work to unify 
attention loss $p_{\text{att}}(C|X)$ and CTC loss $p_{\text{ctc}}(C|X)$ with a linear interpolation weight $\lambda$, as follows:
\begin{equation}
\label{eq:mtl}
\mathcal{L}_{\text{mod}}=\lambda\,\log\,p_{\text{ctc}}(C|X)+(1-\lambda)\,\log\,p_{\text{att}}^{*}(C|X)
\end{equation}
The unified model allows to obtain both monotonicity and effective sequence level training. 

$p_{\text{att}}\left(C|X\right)$ represents the posterior probability of character label sequence $C$ w.r.t input sequence
$X$ based on the attention approach, which is decomposed with the probabilistic chain rule, as follows:
\begin{equation}
\label{eq:att}
p_{\text{att}}^{*}\left(C|X\right)\thickapprox \prod _{l=1} ^L p\left(c_{l}|c_{1}^{*},....,c_{l-1}^{*},\ X\right),
\end{equation}
where $c_{l}^{*}$ denotes the ground truth history.
Detailed explanations about the attention mechanism is described later.

Similarly, $p_{\text{ctc}}\left(C|X\right)$ represents the posterior probability based on the CTC approach.
\begin{equation}
p_{\text{ctc}}\left(C|X\right)\thickapprox \sum _{Z \in \mathcal{Z}(C)} p(Z|X),
\end{equation}
where $Z= (\mathbf{z}_t | t=1, \dots, T)$ is a CTC state sequence composed of the original grapheme set and the additional blank symbol. 
$\mathcal{Z}(C)$ is a set of all possible sequences given the character sequence $C$.

The following paragraphs explain the encoder, attention decoder, CTC, and joint decoding used in our approach.

\subsubsection*{Encoder}
In our approach, both CTC and attention use the same encoder function, as follows:
\begin{equation}
 \label{eq:enc}
 \mathbf{h}_t = \text{Encoder}(X),
\end{equation}
where $\mathbf{h}_t $ is an encoder output state at $t$.
As an encoder function $\text{Encoder}(\cdot)$, we use bidirectional LSTM (BLSTM) or deep CNN followed by BLSTMs.
Convolutional neural networks (CNN) has achieved great success in image recognition \cite{vgg}. 
Previous studies applying CNN in seq2seq speech recognition \cite{vggspeech2} also showed that incorporating
a deep CNNs in the encoder could further boost the performance.

In this work, we investigate the effect of convolutional layers in
joint CTC-attention framework for multilingual setting. We use the
initial 6 layers of VGG net architecture \cite{vgg} in table \ref{tab:Experiment-details}.
For each speech feature image, one feature map is formed initially.
VGG net then extracts 128 feature maps, where each feature map is
downsampled to $(1/4\times1/4)$ images along time-frequency axis
via the two maxpooling layers with $stride=2$. 

\subsubsection*{Attention Decoder:}
Location aware attention mechanism \cite{locatt} is used in this work.
Equation~\eqref{locatt} denotes the output of location aware attention, where $a_{lt}$ acts as an attention weight.
\begin{equation}
\label{locatt}
a_{lt}=\text{LocationAttention}\left(\left\{ a_{l-1}\right\} _{t=1}^{T}, \mathbf{q}_{l-1}, \mathbf{h}_{t}\right).
\end{equation}
Here, $\mathbf{q}_{l-1}$ denotes the decoder hidden state, $\mathbf{h}_{t}$ is the encoder output state as shown in equation~\eqref{eq:enc}.
The location attention function represents a convolution function \textbf{*} as in equation \eqref{conv_feat}. 
It maps the attention weight of the previous label $a_{l-1}$ to a multi channel view $\mathbf{f}_{t}$ for better representation. 
\begin{align}
\mathbf{f}_{t} & =\mathbf{K}*\text{\ensuremath{\mathbf{a}}}_{l-1}, \label{conv_feat} \\
e_{lt} & = \mathbf{g}^{T} \tanh(\text{Lin}(\mathbf{q}_{l-1})+ \text{Lin}(\mathbf{h}_{t})+\text{LinB}(\mathbf{f}_{t})), \label{energy}\\
a_{lt} & = \text{Softmax}(\{e_{lt}\}_{t=1}^{T}) \label{softmax}
\end{align}
Equation~\eqref{energy} provides the unnormalized attention vectors computed with the learnable vector $\mathbf{g}$, linear transformation $\text{Lin}(\cdot)$, and affine transformation $\text{LinB}(\cdot)$.
Equation~\eqref{softmax} computes a normalized attention weight based on the softmax operation $\text{Softmax}(\cdot)$.   
Finally, the context vector $\mathbf{r}_{l}$ is obtained by the weighted summation of the encoder output state $\mathbf{h}_{t}$ over entire frames with the attention weight as follows:
\begin{equation}
\mathbf{r}_{l}=\sum_{t=1}^{T}a_{lt}\mathbf{h}_{t}.
\end{equation}

The decoder function is an LSTM layer which decodes the next character output label $c_{l}$ from their previous label $c_{l-1}$, hidden state of the decoder $q_{l-1}$ and attention output $\mathbf{r}_{l}$, as follows:
\begin{equation}
p\left(c_{l}|c_{1},....,c_{l-1},\ X\right)=\text{Decoder}(\mathbf{r}_{l},q_{l-1},c_{l-1})
\end{equation}
This equation is incrementally applied to form $p_{\text{att}}^{*}$ in equation~\eqref{eq:att}.

\subsubsection*{Connectionist temporal classification (CTC):}
Unlike the attention approach, CTC do not use any specific decoder. 
Instead it invokes two important components to perform character level training and decoding. 
First component, is an RNN based encoding module $p(Z|X)$.
The second component contains a language model and state transition module. 
The CTC formalism is a special case \cite{graves2012supervised,hori2017advances} of hybrid DNN-HMM framework with an inclusion of Bayes rule to obtain $p(C|X)$.

\subsubsection*{Joint decoding:}

Once we have both CTC and attention-based seq2seq models trained, both are jointly used for decoding as below:
\begin{align}
\begin{split}\log p_{\text{hyp}}(c_{l}|c_{1},....,c_{l-1}, X)=\\
\alpha\log p_{\text{ctc}}(c_{l}|c_{1},....,c_{l-1}, X)\\
+(1-\alpha)\log p_{\text{att}}(c_{l}|c_{1},....,c_{l-1}, X)
\end{split}
\label{joint decoding}
\end{align}
Here $\log p_{\text{hyp}}$ is a final score used during beam search.
$\alpha$ controls the weight between attention and CTC models.
$\alpha$ and multi-task learning weight $\lambda$ in equation~\eqref{eq:mtl} are set differently in our experiments.

\begin{table}[tbh]
\caption{\label{tab:Details-of-the}Details of the BABEL data used for performing
the multilingual experiments}
\scalebox{1.15}{%
\centering{}{\tiny{}}%
\begin{tabular}{|c|c|c|c|c|c|c|}
\hline 
\multirow{2}{*}{{\tiny{}Usage}} & \multirow{2}{*}{{\tiny{}Language}} & \multicolumn{2}{c|}{{\tiny{}Train}} & \multicolumn{2}{c|}{{\tiny{}Eval}} & \multirow{2}{*}{{\tiny{}\# of characters}}\tabularnewline
\cline{3-6} 
 &  & {\tiny{}\# spkrs.} & {\tiny{}\# hours} & {\tiny{}\# spkrs.} & {\tiny{}\# hours} & \tabularnewline
\hline 
\hline 
\multirow{10}{*}{{\tiny{}Train}} & {\tiny{}Cantonese } & {\tiny{}952} & {\tiny{}126.73} & {\tiny{}120} & {\tiny{}17.71} & {\tiny{}3302}\tabularnewline
\cline{2-7} 
 & {\tiny{}Bengali } & {\tiny{}720} & {\tiny{}55.18} & {\tiny{}121} & {\tiny{}9.79} & {\tiny{}66}\tabularnewline
\cline{2-7} 
 & {\tiny{}Pashto} & {\tiny{}959} & {\tiny{}70.26} & {\tiny{}121} & {\tiny{}9.95} & {\tiny{}49}\tabularnewline
\cline{2-7} 
 & {\tiny{}Turkish} & {\tiny{}963} & {\tiny{}68.98} & {\tiny{}121} & {\tiny{}9.76} & {\tiny{}66}\tabularnewline
\cline{2-7} 
 & {\tiny{}Vietnamese} & {\tiny{}954} & {\tiny{}78.62} & {\tiny{}120} & {\tiny{}10.9} & {\tiny{}131}\tabularnewline
\cline{2-7} 
 & {\tiny{}Haitian} & {\tiny{}724} & {\tiny{}60.11} & {\tiny{}120} & {\tiny{}10.63} & {\tiny{}60}\tabularnewline
\cline{2-7} 
 & {\tiny{}Tamil } & {\tiny{}724} & {\tiny{}62.11} & {\tiny{}121} & {\tiny{}11.61} & {\tiny{}49}\tabularnewline
\cline{2-7} 
 & {\tiny{}Kurdish} & {\tiny{}502} & {\tiny{}37.69} & {\tiny{}120} & {\tiny{}10.21} & {\tiny{}64}\tabularnewline
\cline{2-7} 
 & {\tiny{}Tokpisin} & {\tiny{}482} & {\tiny{}35.32} & {\tiny{}120} & {\tiny{}9.88} & {\tiny{}55}\tabularnewline
\cline{2-7} 
 & {\tiny{}Georgian} & {\tiny{}490} & {\tiny{}45.35} & {\tiny{}120} & {\tiny{}12.30} & {\tiny{}35}\tabularnewline
\hline 
\multirow{4}{*}{{\tiny{}Target}} & {\tiny{}Assamese} & {\tiny{}720} & {\tiny{}54.35} & {\tiny{}120} & {\tiny{}9.58} & {\tiny{}66}\tabularnewline
\cline{2-7} 
 & {\tiny{}Tagalog} & {\tiny{}966} & {\tiny{}44.0} & {\tiny{}120} & {\tiny{}10.60} & {\tiny{}56}\tabularnewline
\cline{2-7} 
 & {\tiny{}Swahili} & {\tiny{}491} & {\tiny{}40.0} & {\tiny{}120} & {\tiny{}10.58} & {\tiny{}56}\tabularnewline
\cline{2-7} 
 & {\tiny{}Lao} & {\tiny{}733} & {\tiny{}58.79} & {\tiny{}119} & {\tiny{}10.50} & {\tiny{}54}\tabularnewline
\hline 
\end{tabular}{\tiny \par}
}
\end{table}

\begin{table}[tbh]
\caption{Experiment details\label{tab:Experiment-details}}
\begin{centering}
{\scriptsize{}}%
\begin{tabular}{cc}
\hline 
{\scriptsize{}Model Configuration} & \tabularnewline
\hline 
\hline 
{\scriptsize{}Encoder} & {\scriptsize{}Bi-RNN}\tabularnewline
{\scriptsize{}\# encoder layers} & {\scriptsize{}5}\tabularnewline
{\scriptsize{}\# encoder units} & {\scriptsize{}320}\tabularnewline
{\scriptsize{}\# projection units} & {\scriptsize{}320}\tabularnewline
{\scriptsize{}Decoder} & {\scriptsize{}Bi-RNN}\tabularnewline
{\scriptsize{}\# decoder layers} & {\scriptsize{}1}\tabularnewline
{\scriptsize{}\# decoder units} & {\scriptsize{}300}\tabularnewline
{\scriptsize{}\# projection units} & {\scriptsize{}300}\tabularnewline
{\scriptsize{}Attention} & {\scriptsize{}Location-aware}\tabularnewline
{\scriptsize{}\# feature maps} & {\scriptsize{}10}\tabularnewline
{\scriptsize{}\# window size} & {\scriptsize{}100}\tabularnewline
\hline 
\hline 
{\scriptsize{} Training Configuration} & \tabularnewline
\hline 
\hline 
{\scriptsize{}MOL} & {\scriptsize{}$5e^{-1}$}\tabularnewline
{\scriptsize{}Optimizer} & {\scriptsize{}AdaDelta}\tabularnewline
{\scriptsize{}Initial learning rate} & {\scriptsize{}1.0}\tabularnewline
{\scriptsize{}AdaDelta $\epsilon$} & {\scriptsize{}$1e^{-8}$}\tabularnewline
{\scriptsize{}AdaDelta $\epsilon$ decay} & {\scriptsize{}$1e^{-2}$}\tabularnewline
{\scriptsize{}Batch size} & {\scriptsize{}30}\tabularnewline
{\scriptsize{}Optimizer} & {\scriptsize{}AdaDelta}\tabularnewline
\hline 
\hline 
{\scriptsize{}Decoding Configuration} & \tabularnewline
\hline 
\hline 
{\scriptsize{}Beam size} & {\scriptsize{}20}\tabularnewline
{\scriptsize{}ctc-weight} & {\scriptsize{}$3e^{-1}$}\tabularnewline
\hline 
\end{tabular}
\par\end{centering}{\scriptsize \par}
\vspace{0.5cm}
\centering{}\subfloat[Convolutional layers in joint CTC-attention]{
\raggedleft{}{\scriptsize{}}%
\begin{tabular}{cc}
\hline 
\multicolumn{2}{c}{{\scriptsize{}CNN Model Configuration -2 components}}\tabularnewline
\hline 
\hline 
{\scriptsize{}Component 1} & {\scriptsize{}2 convolution layers}\tabularnewline
{\scriptsize{}Convolution 2D} & {\scriptsize{}in = 1, out = 64, filter = 3$\times$ 3}\tabularnewline
{\scriptsize{}Convolution 2D} & {\scriptsize{}in = 64, out = 64, filter = 3$\times$ 3}\tabularnewline
{\scriptsize{}Maxpool 2D} & {\scriptsize{}patch = 2$\times$2, stride = 2$\times$2}\tabularnewline
{\scriptsize{}Component 2} & {\scriptsize{}2 convolution layers}\tabularnewline
{\scriptsize{}Convolution 2D} & {\scriptsize{}in = 64, out = 128, filter = 3$\times$ 3}\tabularnewline
{\scriptsize{}Convolution 2D} & {\scriptsize{}in = 128, out = 128, filter = 3$\times$ 3}\tabularnewline
{\scriptsize{}Maxpool 2D} & {\scriptsize{}patch = 2$\times$2, stride = 2$\times$2}\tabularnewline
\hline 
\end{tabular}{\scriptsize \par}}
\end{table}

\section{\label{sec:Data-and-Experimental}Data details and experimental setup}

In this work, the experiments are conducted using the BABEL speech corpus collected from the IARPA babel program. 
The corpus is mainly composed of conversational telephone speech (CTS) but some scripted recordings and far field recordings are presented as well. 
Table~\ref{tab:Details-of-the} presents the details of the languages used in this work for training and evaluation.
\begin{table*}
\caption{\label{tab:Recognition-performance-of}Recognition performance of
naive multilingual approach for eval set of 10 BABEL training languages
trained with the train set of same languages }
\scalebox{1.15}{%
\centering{}{\scriptsize{}}%
\begin{tabular}{ccccccccccc}
\hline 
{\scriptsize{}\%CER on Eval set} & \multicolumn{10}{c}{{\scriptsize{}Target languages}}\tabularnewline
\cline{2-11} 
{\scriptsize{}for} & {\scriptsize{}Bengali} & {\scriptsize{}Cantonese} & {\scriptsize{}Georgian} & {\scriptsize{}Haitian} & {\scriptsize{}Kurmanji} & {\scriptsize{}Pashto} & {\scriptsize{}Tamil} & {\scriptsize{}Turkish} & {\scriptsize{}Tokpisin} & {\scriptsize{}Vietnamese}\tabularnewline
\hline 
{\scriptsize{}Monolingual - BLSTMP} & {\scriptsize{}43.4} & {\scriptsize{}37.4} & {\scriptsize{}35.4} & {\scriptsize{}39.7} & {\scriptsize{}55.0} & {\scriptsize{}37.3} & {\scriptsize{}55.3} & {\scriptsize{}50.3} & {\scriptsize{}32.7} & {\scriptsize{}54.3}\tabularnewline
{\scriptsize{}Multilingual - BLSTMP} & {\scriptsize{}42.9 } & {\scriptsize{}36.3 } & {\scriptsize{}38.9} & {\scriptsize{}38.5} & {\scriptsize{}52.1} & {\scriptsize{}39.0} & {\scriptsize{}48.5} & {\scriptsize{}36.4} & {\scriptsize{}31.7} & {\scriptsize{}41.0}\tabularnewline
{\scriptsize{}+ VGG} & {\scriptsize{}39.6} & {\scriptsize{}34.3} & {\scriptsize{}36.0} & {\scriptsize{}34.5} & {\scriptsize{}49.9} & {\scriptsize{}34.7} & {\scriptsize{}45.5} & {\scriptsize{}28.7} & {\scriptsize{}33.7} & {\scriptsize{}37.4}\tabularnewline
\hline 
\end{tabular}{\scriptsize \par}
}
\end{table*}

80 dimensional Mel-filterbank (fbank) features are then extracted from the speech samples using a sliding window of size 25 ms with 10ms stride. 
KALDI toolkit \cite{povey2011kaldi} is used to perform
the feature processing. 
The fbank features are then fed to a seq2seq model with the following configuration:

The Bi-RNN \cite{schuster1997bidirectional} models mentioned above uses a LSTM \cite{hochreiter1997long} cell followed by a projection layer (BLSTMP). 
In our experiments below, we use only a character-level seq2seq model trained by CTC and attention decoder. 
Thus in the following experiments we intend to use character error rate (\% CER) as a suitable measure to analyze the model performance. 
However, in section \ref{subsec:Multilingual-RNNLM} we integrate a character-level RNNLM \cite{mikolov2011rnnlm} with seq2seq model externally and showcase the performance in terms of word error rate (\% WER). 
In this case the words are obtained by concatenating the characters and the space together for scoring with reference words.
All experiments are implemented in ESPnet, end-to-end speech processing toolkit \cite{watanabe2018espnet}.

\section{Multilingual experiments}\label{sec:Multilingual-experiments}

Multilingual approaches used in hybrid RNN/DNN-HMM systems~\cite{karafiat2016multilingual} have been
used for for tackling the problem of low-resource data condition.
Some of these approaches include language adaptive training and shared
layer retraining \cite{tong2017investigation}. Among them, the most
benefited method is the parameter sharing technique \cite{karafiat2016multilingual}.
To incorporate the former approach into encoder, CTC and attention
decoder model, we performed the following experiments:
\begin{itemize}
\item Stage 0 - Naive training combining all languages 
\item Stage 1 - Retraining the decoder (both CTC and attention) after 
initializing with the multilingual model from stage-0
\item Stage 2 - The resulting model obtained from stage-1 is further retrained
across both encoder and decoder

\begin{table}[H]
\caption{Comparison of naive approach and training only the last layer performed
using the Assamese language\label{tab:Comparison-of-naive} }
\centering{}{\scriptsize{}}%
\begin{tabular}{cccc}
\hline 
{\scriptsize{}Model type} & {\scriptsize{}Retraining } & {\scriptsize{}\% CER} & {\scriptsize{}\% Absolute gain}\tabularnewline
\hline 
\hline 
{\scriptsize{}Monolingual} & {\scriptsize{}-} & {\scriptsize{}45.6} & {\scriptsize{}-}\tabularnewline
{\scriptsize{}Multi. (after $4^{th}$ epoch)} & {\scriptsize{}Stage 1} & {\scriptsize{}61.3} & {\scriptsize{}-15.7}\tabularnewline
{\scriptsize{}Multi. (after $4^{th}$ epoch)} & {\scriptsize{}Stage 2} & {\scriptsize{}44.0} & {\scriptsize{}1.6}\tabularnewline
{\scriptsize{}Multi. (after $15^{th}$ epoch)} & {\scriptsize{}Stage 2} & {\scriptsize{}41.3} & {\scriptsize{}4.3}\tabularnewline
\hline 
\end{tabular}{\scriptsize \par}
\end{table}
\end{itemize}

\subsection{\label{subsec:Naive-approach-}Stage 0 - Naive approach}

In this approach, the model is first trained with 10 multiple languages
as denoted in table \ref{tab:Details-of-the} approximating to 600
hours of training data. data from all languages available during training
is used to build a single seq2seq model. The model is trained with a 
character label set composed of characters from all languages including both train and target
set as mentioned in table \ref{tab:Details-of-the}. The model provides better
generalization across languages. Languages with limited data when
trained with other languages allows them to be robust and helps in
improving the recognition performance. In spite of being simple, the
model has limitations in keeping the target language data unseen during
training.
\subsubsection*{Comparison of VGG-BLSTM and BLSTMP}
Table \ref{tab:Recognition-performance-of} shows the recognition
performance of naive multilingual approach using BLSTMP and VGG model
against a monolingual model trained with BLSTMP. The results clearly
indicate that having a better architecture such as VGG-BLSTM helps
in improving multilingual performance. Except Pashto, Georgian and
Tokpisin, the multilingual VGG-BLSTM model gave 8.8 \% absolute gain
in average over monolingual model. In case of multilingual BLSTMP,
except Pashto and Georgian an absolute gain of 5.0 \% in average
is observed over monolingual model. Even though the VGG-BLSTM gave
improvements, we were not able to perform stage-1 and stage-2 retraining
with it due to time constraints. Thus, we proceed further with multilingual
BLSTMP model for retraining experiments tabulated below.

\subsection{\label{subsec:retrain}Stage 1 - Retraining decoder only}

To alleviate the limitation in the previous approach, the final layer
of the seq2seq model which is mainly responsible for classification
is retrained to the target language. 
\begin{figure}[H]
\includegraphics[scale=0.55]{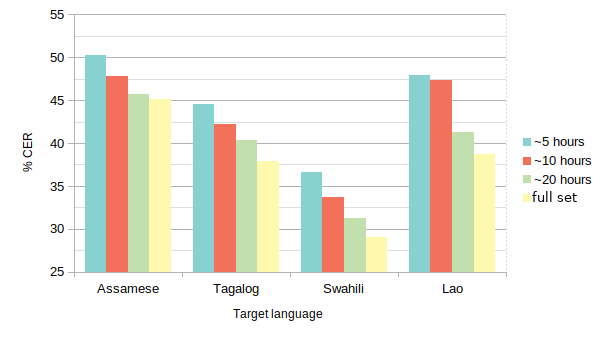}
\caption{Difference in performance for 5 hours, 10 hours, 20 hours and full set \label{fig:Difference-in-performance}of
target language data used to retrain a multilingual model from stage-1}
\end{figure}
In previous works \cite{karafiat2016multilingual,tong2017investigation}
related to hybrid DNN/RNN models and CTC based models \cite{tong2017multilingual,dalmia2018sequence}
the softmax layer is only adapted. However in our case, the attention
decoder and CTC decoder both have to be retrained to the target language.
This means the CTC and attention layers are only updated for gradients
during this stage.  We found using SGD optimizer with initial learning
rate of $1e^{-4}$ works better for retraining compared to AdaDelta.

The learning rate is decayed in this training at a factor of $1e^{-1}$
if there is a drop in validation accuracy. Table \ref{tab:Comparison-of-naive}
shows the performance of simply retraining the last layer using a
single target language Assamese.

\subsection{Stage 2 - Finetuning both encoder and decoder}\label{subsec:finetune}

Based on the observations from stage-1 model in section \ref{subsec:retrain},
we found that simply retraining the decoder towards a target language
resulted in degrading \%CER the performance from 45.6 to 61.3. This
is mainly due to the difference in distribution across encoder and
decoder. So, to alleviate this difference the encoder and decoder
is once again retrained or fine-tuned using the model from stage-1.
The optimizer used here is SGD as in stage-1, but the initial learning
rate is kept to $1e^{-2}$ and decayed based on validation performance.
The resulting model gave an absolute gain of 1.6\% when finetuned
a multilingual model after 4th epoch. Also, finetuning a model after
15th epoch gave an absolute gain of 4.3\%.
\begin{table}[H]
\caption{Stage-2 retraining across all languages with full set of target language
data\label{tab:Fine-tuning-across-all}}
\centering{}{\scriptsize{}}%
\begin{tabular}{ccccc}
\hline 
{\scriptsize{}\% CER on} & \multicolumn{3}{c}{{\scriptsize{}Target Languages}} & \tabularnewline
\cline{2-5} 
{\scriptsize{}eval set} & {\scriptsize{}Assamese} & {\scriptsize{}Tagalog} & {\scriptsize{}Swahili} & {\scriptsize{}Lao}\tabularnewline
\hline 
\hline 
{\scriptsize{}Monolingual} & {\scriptsize{}45.6} & {\scriptsize{}43.1} & {\scriptsize{}33.1} & {\scriptsize{}42.1}\tabularnewline
{\scriptsize{}Stage-2 retraining} & {\scriptsize{}41.3} & {\scriptsize{}37.9} & {\scriptsize{}29.1} & {\scriptsize{}38.7}\tabularnewline
\hline 
\end{tabular}{\scriptsize \par}
\end{table}
To further investigate the performance of this approach across different
target data sizes, we split the train set into $\sim$5 hours, $\sim$10
hours, $\sim$20 hours and $\sim$full set. Since, in this approach
the model is only finetuned by initializing from stage-1 model, the
model architecture is fixed for all data sizes. Figure \ref{fig:Difference-in-performance}
shows the effectiveness of finetuning both encoder and decoder. The
gains from 5 to 10 hours was more compared to 20 hours to full set. 

Table \ref{tab:Fine-tuning-across-all} tabulates the \% CER obtained
by retraining the stage-1 model with $\sim$full set of target language
data. An absolute gain is observed using stage-2 retraining across
all languages compared to monolingual model.




\subsection{Multilingual RNNLM\label{subsec:Multilingual-RNNLM}}

In an ASR system, a language model (LM) takes an important role by incorporating
external knowledge into the system. Conventional ASR systems combine
an LM with an acoustic model by FST giving a huge performance gain. This
trend is also shown in general including hybrid ASR systems and neural network-based sequence-to-sequence ASR systems. 

The following experiments show a benefit of using a language model in decoding with the previous stage-2 transferred models. Although the performance gains in \%CER are also generally observed  over all target languages, the improvement in \%WER was more distinctive. The results shown in the following Fig. \ref{fig:RNNLM-plot} are in \%WER. 
``whole'' in each figure means we used all the available data for the target language as full set explained before.
\vspace{-0.32cm}
\begin{figure}[H]
\centering
\includegraphics[scale=0.45]{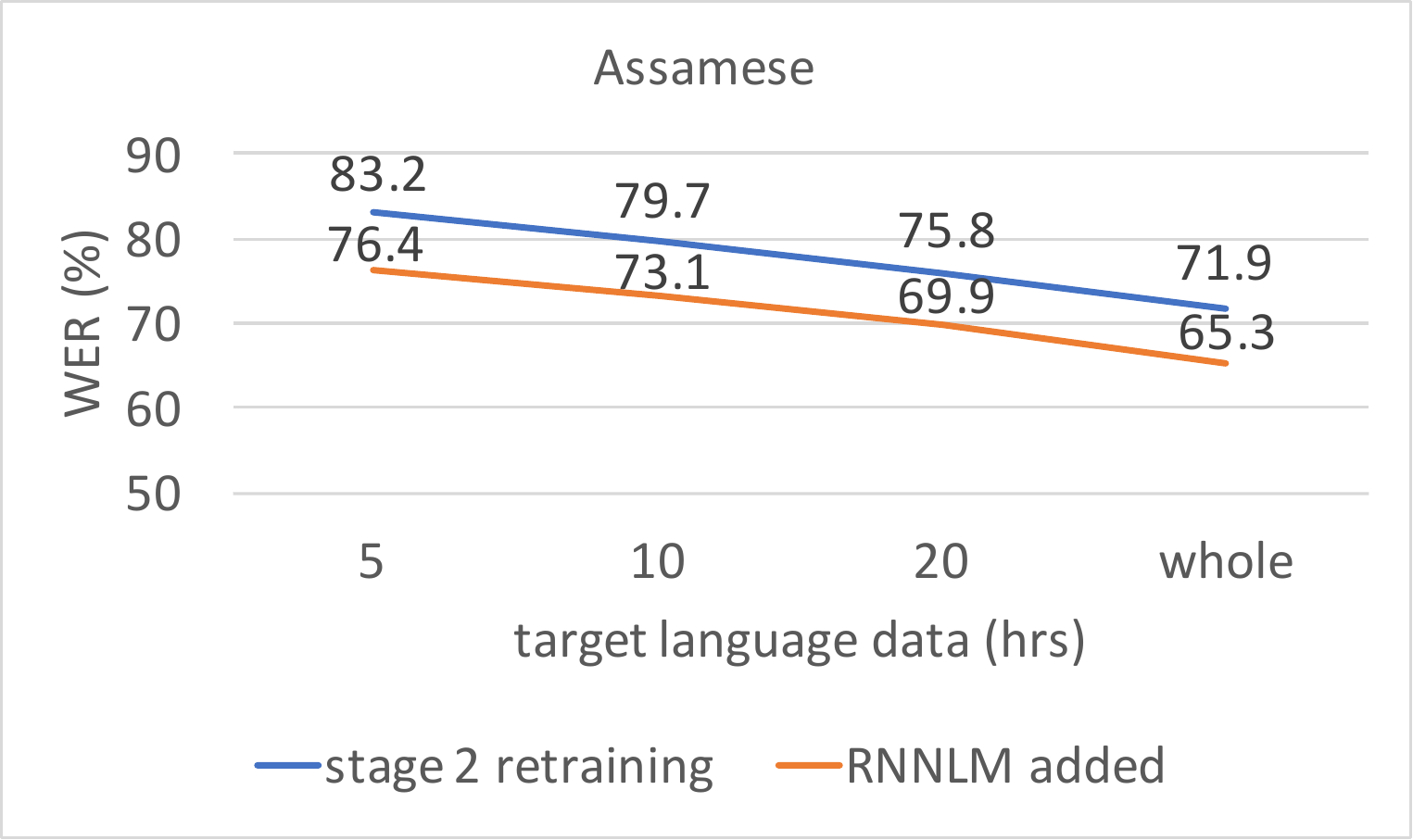}

\includegraphics[scale=0.45]{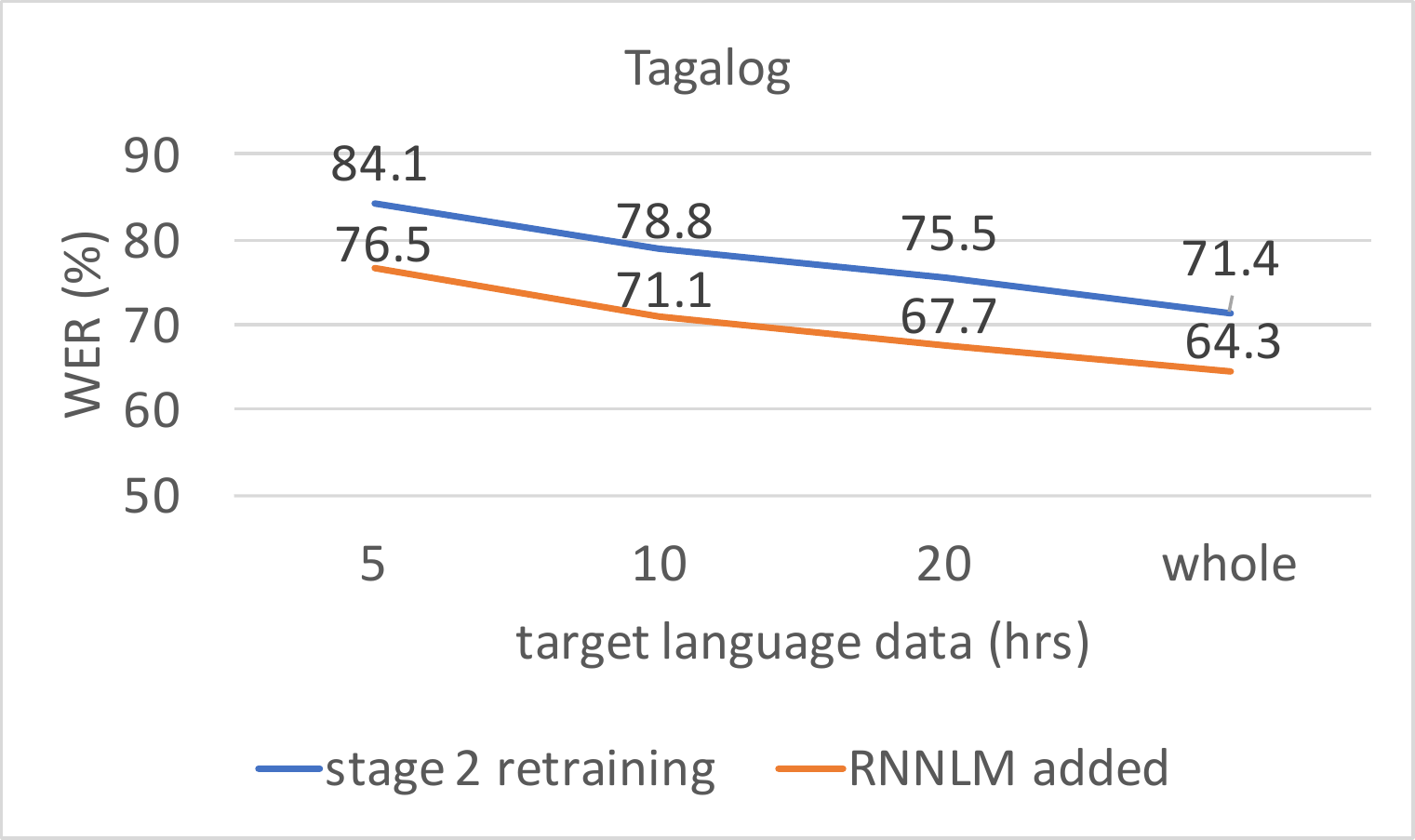}

\includegraphics[scale=0.45]{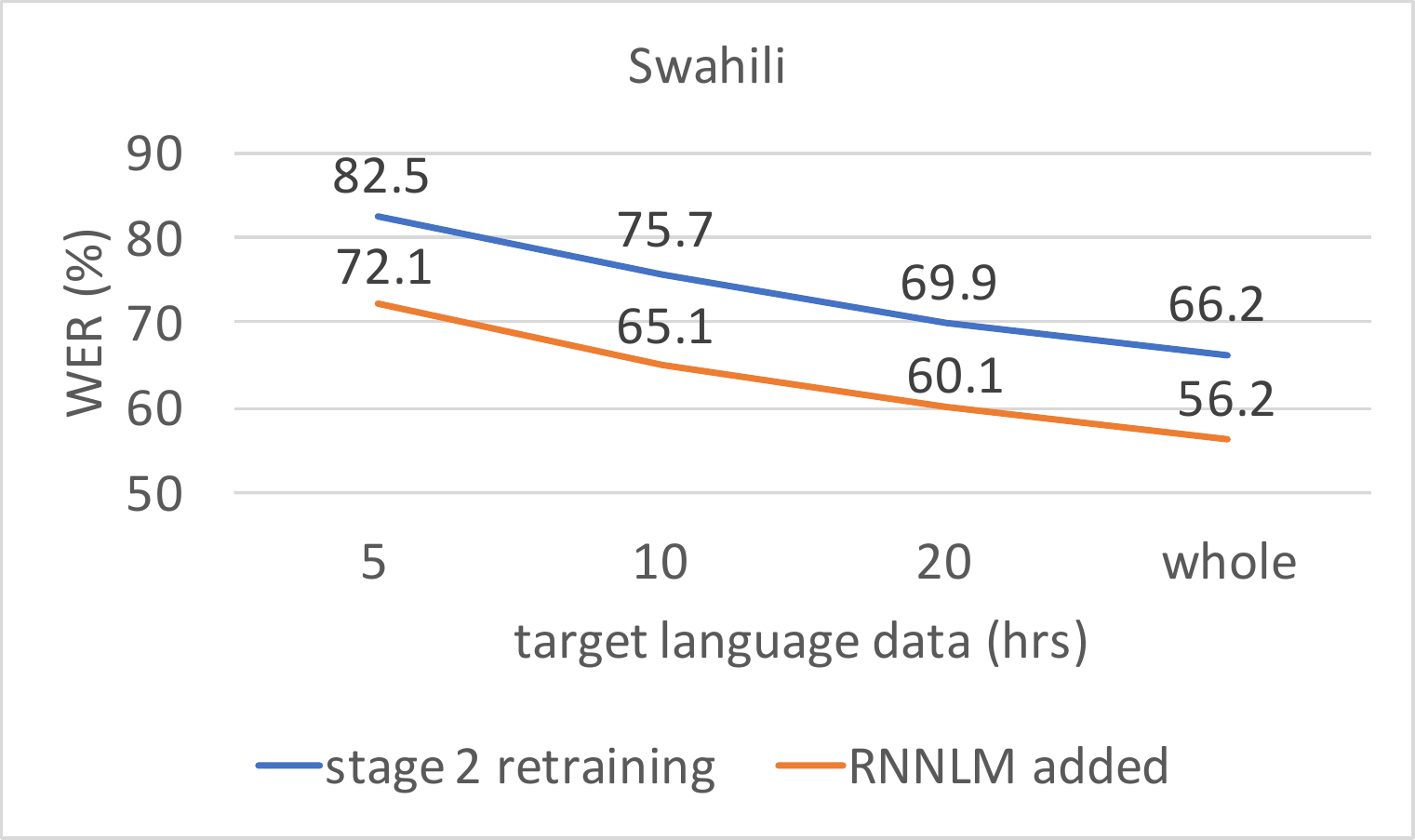}

\includegraphics[scale=0.45]{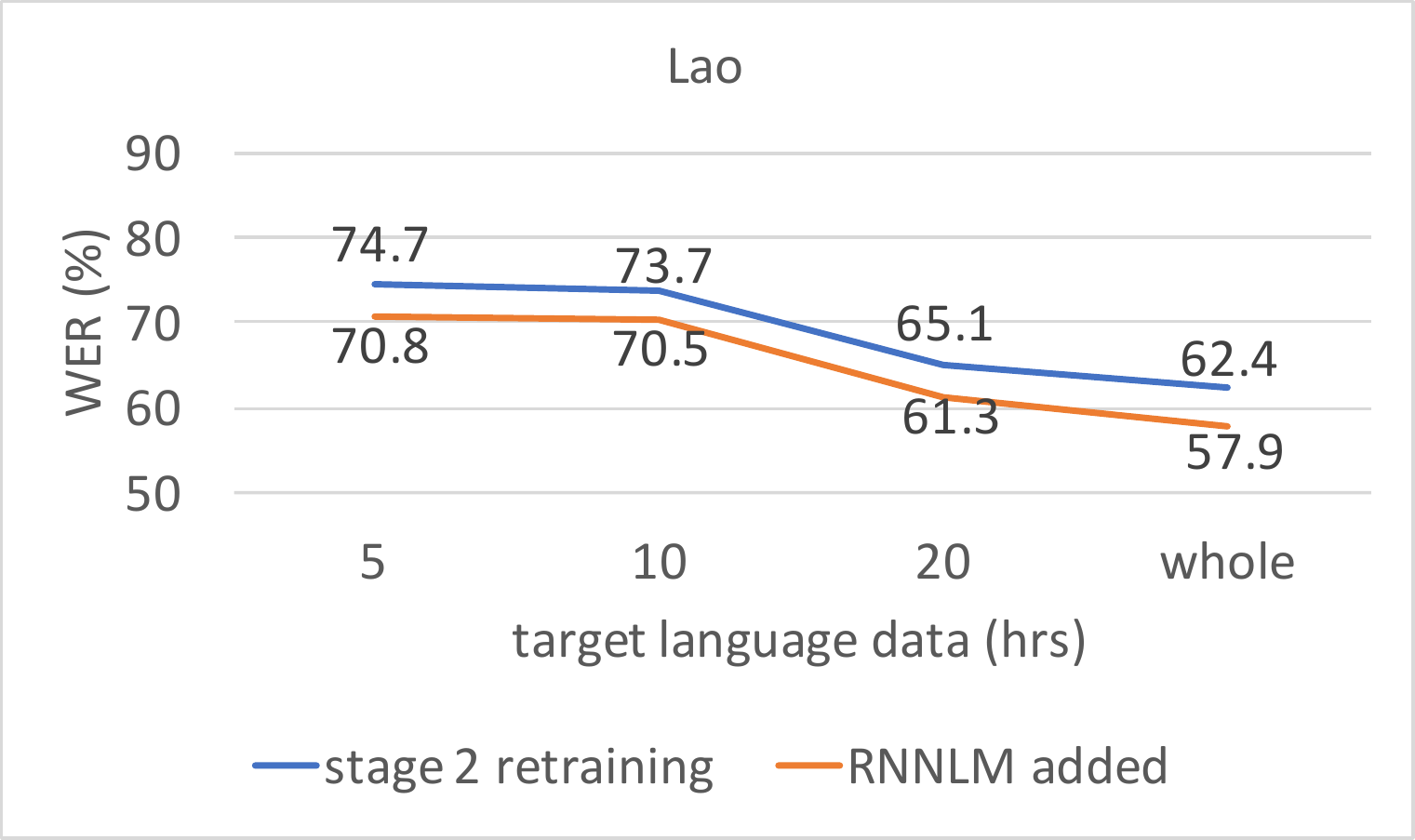}

\caption{Recognition performance after integrating RNNLM during decoding in \%WER for different amounts of target data\label{fig:RNNLM-plot}}
\end{figure}
\vspace{-0.34cm}

We used a character-level RNNLM, which was trained with 2-layer LSTM on character sequences. We use all available paired text in the corresponding target language to train the LM for the language. No external text data were used. All language models are trained separately from the seq2seq models. When building dictionary, we combined all the characters over all 15 languages
mentioned in table \ref{tab:Details-of-the} 
to make them work with transferred models. Regardless of the amount of data used for transfer learning, the RNNLM provides consistent gains across all languages over different 
data sizes.

{\scriptsize{}}
\begin{table}[H]
{\scriptsize{}\caption{\label{tab:Recognition-performance-interms}Recognition performance
in \%WER using stage-2 retraining and multilingual RNNLM}
}{\scriptsize \par}
\centering{}{\scriptsize{}}%
\begin{tabular}{ccccc}
\hline 
\multirow{2}{*}{{\scriptsize{}Model type}} & \multicolumn{4}{c}{{\scriptsize{}\%WER on target languages}}\tabularnewline
\cline{2-5} 
 & \multicolumn{1}{c}{{\scriptsize{}Assamese}} & \multicolumn{1}{c}{{\scriptsize{}Tagalog}} & \multicolumn{1}{c}{{\scriptsize{}Swahili}} & \multicolumn{1}{c}{{\scriptsize{}Lao}}\tabularnewline
\hline 
{\scriptsize{}Stage-2 retraining} & {\scriptsize{}71.9} & {\scriptsize{}71.4} & {\scriptsize{}66.2} & {\scriptsize{}62.4}\tabularnewline
{\scriptsize{}+ Multi. RNNLM} & {\scriptsize{}65.3} & {\scriptsize{}64.3} & {\scriptsize{}56.2} & {\scriptsize{}57.9}\tabularnewline
\hline 
\end{tabular}{\scriptsize \par}
\end{table}
{\scriptsize \par}

As explained already, language models were trained separately and
used to decode jointly with seq2seq models. The intuition behind it
is to use the separately trained language model as a complementary
component that works with a implicit language model within a seq2seq
decoder. The way of RNNLM assisting decoding follows the equation below:

\begin{align}
\begin{split}\log p(c_{l}|c_{1:l-1}, X)=&
\log p_{\text{hyp}}(c_{l}|c_{1:l-1}, X)\\
&+\beta\log p_{\text{lm}}(c_{l}|c_{1:l-1}, X)
\end{split}
\label{decoding with rnnlm}
\end{align}
$\beta$ is a scaling factor that combines the scores from a joint decoding eq.(\ref{joint decoding}) with RNN-LM, denoted as $p_{\text{lm}}$. This approach is called shallow fusion.

Our experiments for target languages show that the gains from adding
RNNLM are consistent regardless of the amount of data used
for transfer learning. In other words, in Figure~\ref{fig:RNNLM-plot}, the gap between two lines are almost consistent over all languages. 

Also, we observe the gain we get by adding RNN-LM in decoding is large. For example, in the case of assamese, the gain by RNN-LM in decoding with a model retrained on 5
hours of the target language data is almost comparable
with the model stage-2 retrained with 20 hours of target language data. On average, absolute gain $\sim$6\% is obtained across all target
languages as noted in table \ref{tab:Recognition-performance-interms}.

\section{Conclusion}

In this work, we have shown the importance of transfer learning approach such as stage-2 multilingual retraining in a seq2seq model setting. Also, careful selection of train and target languages from BABEL provide a wide variety
in recognition performance (\%CER) and helps in understanding the efficacy
of seq2seq model. The experiments using character-based RNNLM showed the importance of language model in boosting recognition performance (\%WER) over all different hours of target data available for transfer learning.

Table \ref{tab:Fine-tuning-across-all} and \ref{tab:Recognition-performance-interms} summarizes, the effect of these techniques in terms of \%CER and \%WER. These methods also show their flexibility in incorporating it in attention and CTC based seq2seq model without compromising loss in performance.

\section{Future work}

We could use better architectures such as VGG-BLSTM as a multilingual prior
model before transferring them to a new target language by performing
stage-2 retraining. The naive multilingual approach can be improved
by including language vectors as input or target during training to
reduce the confusions. Also, investigation of multilingual bottleneck
features \cite{grezl2014adaptation} for seq2seq model can provide
better performance. Apart from using the character level language
model as in this work, a word level RNNLM can be connected during
decoding to further improve \%WER. The attention based decoder can
be aided with the help of RNNLM using cold fusion approach during
training to attain a better-trained model. In near future, we will incorporate
all the above techniques to get comparable performance with the state-of-the-art
hybrid DNN/RNN-HMM systems.

\bibliographystyle{IEEEbib}
\bibliography{IEEEfull,ref_new}

\end{document}